\documentclass[sigconf]{acmart}

\AtBeginDocument{%
  \providecommand\BibTeX{{%
    \normalfont B\kern-0.5em{\scshape i\kern-0.25em b}\kern-0.8em\TeX}}}

\usepackage{booktabs}
\usepackage{subcaption}
\usepackage{multirow}
\usepackage{pgfplots}

\copyrightyear{2021} 
\acmYear{2021} 
\setcopyright{acmcopyright}\acmConference[CODS COMAD 2021]{8th ACM IKDD CODS and 26th COMAD}{January 2--4, 2021}{Bangalore, India}
\acmBooktitle{8th ACM IKDD CODS and 26th COMAD (CODS COMAD 2021), January 2--4, 2021, Bangalore, India}
\acmPrice{15.00}
\acmDOI{10.1145/3430984.3431030}
\acmISBN{978-1-4503-8817-7/21/01}

\usepackage{graphicx}
\graphicspath{./}

\begin{document}

\title{On-Device Document Classification using multimodal features}

%%\iffalse

\author{Sugam Garg}
\affiliation{Samsung R\&D Institute India-Bangalore}
\email{sugam.g@samsung.com}

\author{Harichandana}
\affiliation{Samsung R\&D Institute India-Bangalore}
\email{hari.ss@samsung.com}

\author{Sumit Kumar}
\affiliation{Samsung R\&D Institute India-Bangalore}
\email{sumit.kr@samsung.com}

%%\fi
\renewcommand{\shortauthors}{Sugam, et al.}

\begin{abstract}
 From small screenshots to large videos, documents take up a bulk of space in a modern smartphone. Documents in a phone can accumulate from various sources, and with the high storage capacity of mobiles, hundreds of documents are accumulated in a short period. However, searching or managing documents remains an onerous task, since most search methods depend on meta-information or only text in a document. In this paper, we showcase that a single modality is insufficient for classification and present a novel pipeline to classify documents on-device, thus preventing any private user data transfer to server. For this task, we integrate an open-source library for Optical Character Recognition (OCR) and our novel model architecture in the pipeline. We optimise the model for size, a necessary metric for on-device inference. We benchmark our classification model with a standard multimodal dataset FOOD-101 and showcase competitive results with the previous State of the Art with 30\% model compression.
\end{abstract}


%% \ccsdesc[500]{Machine Learning~Deep Learning}
%% \ccsdesc[300]{Machine Learning~Image Classification}
%% \ccsdesc{Machine Learning~Text Classification}



\maketitle

\section{Introduction}

With the advent of smartphones with internal memory in GBs, there is a plethora of documents, which can be present on a mobile phone. Some of these are private while some are downloaded just while browsing the internet. Existing search mechanisms heavily rely on the name of the document, which can be random and may not represent the content of the document properly. Thus, an important document may get lost in the clutter, providing a bad user experience. The automatic organization of documents based on its content will immensely increase a user's satisfaction. Since, the contents of a document in a smart phone are personal, sending the document or its content to a server for such kind of processing may lead to privacy and latency issue. Hence, in this paper, we present a light-weight architecture to classify documents on-device, that improves user experience as well as preserve privacy.

Traditionally, document classification considers only text content of a document. The words from a document are converted into vectors, and these vectors are used to compute sentence as well as document vector representations. Sparse Composite Document Vector (SCDV) \cite{mekala2016scdv} calculates document vectors using soft clustering over word vectors. One popular model, Hierarchical Attention Network (HAN), uses a word- and sentence-level attention in classifying documents \cite{yang2016hierarchical}. \cite{adhikari2019rethinking} stipulate that a simple BiLSTM architecture with appropriate regularization yields competitive accuracy and F1-score. \cite{adhikari2019docbert} established the state of the art results for document classification by fine-tuning BERT \cite{devlin2018bert} and demonstrated that BERT could be distilled into a much simpler single-layered lightweight BiLSTM model that provides competitive accuracy. A recently published paper \cite{abreu2019hierarchical}, proposed an approach (HAHNN) takes into account the text structure as well in a document. However, all these models are huge, often containing hundreds of millions of parameters, making on-device deployment infeasible.

Most of the work on document classification is based on text extracted from the document. However, we believe that the organization of the text in the document is also an essential factor for document classification. For example, a boarding pass may be easily identified with the structuring of its data, such as the placement of passenger name, gate number, etc. without having to read the actual text. In this paper, we also consider the organization of text in the virtual space of the document as a feature in form of an image. For this task, we create an in-house dataset of documents with human annotated class category. To validate the efficacy of our multimodal approach, we present results on an open-source multimodal dataset. For on-device, we develop a quantized version of our model. The main contributions of our work are as follows:
\begin{itemize}
 \item Developed a novel multimodal architecture, which considers visual and text features as input for on-device classification.
\item Evaluation on a popular text and image dataset, as there is no standard dataset for on-device document classification.
\item A novel pipeline for on-device document classification that takes input, a pdf document and gives its class as output.
\end{itemize}

\begin{table*}[t] \centering
\begin{tabular}{@{}llllllll@{}}
\toprule
 & \textbf{Modality} & \multicolumn{4}{c}{\textbf{Model Details}} & \textbf{\begin{tabular}[c]{@{}l@{}}Accuracy\end{tabular}} & \textbf{\begin{tabular}[c]{@{}l@{}}Size (MB)\end{tabular}} \\ \midrule
\multirow{8}{*}{\textbf{Baselines}} & \multirow{4}{*}{\textbf{Text}} & \textbf{SVD} & \textbf{1st Layer} & \textbf{2nd Layer} & \textbf{3rd Layer} &  &  \\
 &  & Yes & 2000 & 2000 & 500 & 85.39 & 40 \\
 &  & Yes & 2000 & 1000 & 500 & 85.33 & 20 \\
 &  & No & 2000 & 1000 & 500 & 86.7 & 946 \\ 
 & \multirow{4}{*}{\textbf{Image}} & \multicolumn{2}{l}{\textbf{Batch Norm}} & \textbf{Dropout} & \textbf{Layers Frozen} &  &  \\
 &  & \multicolumn{2}{l}{No} & Yes & 53 & 65.6 & 17 \\
 &  & \multicolumn{2}{l}{Yes} & Yes & 53 & 66.1 & 17 \\
 &  & \multicolumn{2}{l}{Yes} & Yes & 31 & 65.76 & 17 \\ \hline
\multirow{2}{*}{\textbf{Previous Work}} & \multirow{2}{*}{\textbf{\begin{tabular}[c]{@{}l@{}}Text +\\ Image\end{tabular}}} & \multicolumn{4}{l}{Wang et al. 2015 \cite{wang2015recipe}}  & 85.1 & \textgreater 534 \\
 &  & \multicolumn{4}{l}{Kiela et al. 2018 \cite{kiela2018efficient}} & 90.8 & \textgreater 230 \\ \hline
\multirow{4}{*}{\textbf{Fusion}} & \multirow{4}{*}{\textbf{\begin{tabular}[c]{@{}l@{}}Text +\\ Image\end{tabular}}} & \multicolumn{4}{l}{Max} & 86.18 & 12 \\
 &  & \multicolumn{4}{l}{Concatenate} & 89.8 & 13 \\
 &  & \multicolumn{4}{l}{Average} & 82.9 & 12 \\
 &  & \multicolumn{4}{l}{Highway} & 88.03 & 15 \\ \cmidrule(l){1-8} 
\end{tabular}
\caption{Accuracy and model size (in MegaBytes) of fusion models compared to baselines, previous works. (Exact model details of previous works are unknown, thus providing minimum model size through visual model used by each.)}
\label{tab:results}
\vspace{-6mm}
\end{table*}

\section{Background}

Multimodal learning brings out some unique challenges for researchers, given the heterogenity of data. \cite{baltruvsaitis2018multimodal} captures the challenges, methods, and applications of multimodal learning. Document classification is a subjective problem where the classes and data depend on the usecase being targeted. \cite{audebert2019multimodal} classified documents of type questionnaire, memo, etc. and showcased that integrating an additional modality offer more robust representation. They used tesseract-OCR to extract text and generate document embedding, and MobileNetv2 to learn visual features. This approach showed a boost in pure image accuracy by 3\% on Tobacco3482 and RVL-CDIP datasets. In 2015, a new image and text dataset, UPMCFood-101 dataset, with 100K images and 101 classes was proposed by \cite{wang2015recipe}. The researchers built a search engine that retrieves the relevant recipes given an image by using both text and visual features. \cite{kiela2018efficient} verified the performance of multimodal methods on large datasets, and compared various fusion methods with their own method of discretizing continuous features obtained from visual representations. This method demonstrated the feasibility of multimodal methods on large datasets and results showed that multimodal models outperform Fast-Text \cite{joulin2016bag} and the continuous-only approach regardless of the type of fusion. To the best of our knowledge, our document class types have not been used in any document classification method. Amongst, all the possible ways to fuse and co-learn different modalities representation, we choose late fusion for our problem. We want individual modalities to also be able to classify documents, incase the device constrains don't permit a full multimodal classifier. Moreover, it's hard to see low level interactions between visual and text modality in a document, as image of a document hardly describes the textual content of the document.

\section{Approach}

On-device document classification is at a nascent stage, where previous work is sparse. Due to the subjectivity of our task, it is difficult to create a standard dataset that suits the need of all. Thus, to tackle this lack of dataset, we create a small dataset consisting of 5 classes, decided using an internal survey. But, to truly check the efficacy of our model, we needed a dataset, which contains multimodal features. For this, we chose the FOOD-101\cite{wang2015recipe} dataset that contains recipes and images of 101 popular food categories. We use this dataset to benchmark our model\'s performance and to showcase that our multimodal architecture learns both modalities representation and present our experiments below.

\subsection{Baselines}

{\bf Text} The text input of FOOD-101 dataset is the food recipe. We pre-process this text input by performing stop words removal and lemmatization using NLTK \cite{loper2002nltk} Porter Stemmer algorithm. Further, we remove high-frequency(greater than 100,000 occurences) and low-frequency(less than 5 occurences to account for spelling/parsing errors) words from the text since the food category of an recipe is likely to be determined by rare words. Moreover, the maximum number of words in a recipe after eliminating stop words were roughly 100,000. Due to this huge size, it was not practical to train a sequential neural model such as CNN, and RNN for building an on-device classifier since the time complexity of such models is directly proportional to number of words in a sequence. Thus, we use Tf-Idf as feature vectors, and train two fully connected layers and a softmax layer on top of this. To identify recipe category, the order of the recipe is rarely useful, thus the loss of sequential information due to Tf-Idf vectors wouldn't hamper model performance. But the dimensionality of these Tf-Idf vectors is dependent on vocabulary size, and thus it rendered a model of size 720 MB, since the number of parameters first fully connected (FC) layer is directly proportional to the size of input vector, i.e. vocabulary size. Tf-Idf vectors are often sparse and low-rank. So, to resolve this, we use Truncated-SVD to reduce the rank of these high dimensional sparse vectors, and train our classifier on these low-dimensional vectors. We demonstrate in Table \ref{tab:results} that our Tf-Idf vectors were indeed low rank and that with SVD, the model gives an accuracy of ~85\%, a ~1.5\% reduction from the Tf-Idf model but with a reduction in the model size of more than 95\%, since the size of input vector to first FC layer has gone from size of vocab to rank of SVD output.

\begin{figure}[H]
  \centering
  \includegraphics[width=\linewidth]{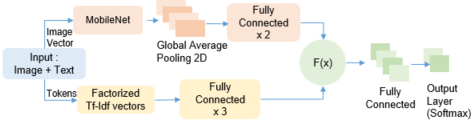}
  \caption{Fusion Model Architecture}
  \Description{Model Architecture}
\label{Model-arch}
\vspace {-4mm}
\end{figure}

{\bf Image} We use a pre-trained MobileNet \cite{howard2017mobilenets} to train a classifier for visual features. MobileNet serves as an ideal choice for on-device classifier as it is optimized for both space and latency. We use transfer learning to retrain a pre-trained MobileNet on our dataset. Our image classifier consists of MobileNet, a pooling layer, two dense layers and a final softmax layer. We freeze the training of MobileNet parameters and train only the final layers for first 15 epochs of training. We treat the number of layers unfrozen after 15th epoch as a hyperparameter. We use batch normalization \cite{ioffe2015batch} and dropout \cite{srivastava2014dropout} to improve the generalization and performance of our model.

\subsection {Fusion Classifier}

We built upon the late fusion strategy of \cite{wang2015recipe} and show improved accuracy with significant compression in the model size. We use our pre-trained MobileNet based image classifier and the text classifer to train a fusion classifier. We transfer the pre-softmax layer features of both the networks and merge the features from both modalities. We train only the layers succeding these merged features and build a classifier as shown in Figure \ref{Model-arch}. We use different methods, F(x), of merging these features, which we discuss below: 

{\bf Concatenation}: We concatenate the vectors from both modalities and train a dense layer on top of the concatenated vector, i.e., 
\begin{equation}
	o(x_n) = W(Ux^t_n \oplus Vx^v_n),
\end{equation}
where W, U, V are the weight matrices of dense layer and $x^{\textit{t}}_{\textit{n}}$ and $x^{\textit{v}}_{\textit{n}}$ are the pre-softmax text and visual features representation respectively. 

{\bf Average}: Here, we retain the softmax layers of the pre-trained models. We merge the output of the softmax layers using component-wise average and train a dense layer on that average, i.e., 
\begin{equation}
	o(x_n) = W(avg(softmax(x^t_n),  softmax( Vx^v_n))),
\end{equation}

{\bf Max}: We combine the information from both modalities using component-wise maximum.
\begin{equation}
	o(x_n) = W max(Ux^t_n, Vx^v_n),
\end{equation}

{\bf Gating Layer post Concatenation}: We concatenate the features of both modalities and train a highway layer \cite{srivastava2015highway} on top of it. 
\begin {equation}
\begin{alignedat}{2}
	y_n = Ux^t_n \oplus Vx^v_n \qquad
	g = Wy_n + b,\\
	t = sigmoid(g) \qquad
	o(x_n) = t * g + (1 - t) * y_n,
\end{alignedat}
\end{equation}

We use ReLU \cite{dahl2013improving} as our non-linear function after merging the layers. The accuracy and model sizes of different fusion strategies as compared to baselines and previous works are shown in Table \ref{tab:results}. With our concatenation model, we were able to match the performance achieved by the Gated model of \cite{kiela2018efficient} with a reduction in model size, as they use a 152-layer Resnet \cite{he2016deep} for capturing visual features while we build upon mobilenet, which has fewer parameters compared to Resnet.

We observed a significant improvement in classification accuracy of food images in the multimodal classifier as compared to individual modality classifiers. For specific classes, as shown in Table \ref{tab:class-analysis}, we observed improvement in classification accuracy with a fusion of image and text features.

\begin{table}[]
\begin{tabular}{@{}lccc@{}}
\toprule
\textbf{Class} & \multicolumn{1}{l}{\textbf{Text}} & \multicolumn{1}{l}{\textbf{Image}} & \multicolumn{1}{l}{\textbf{Fusion}} \\ \midrule
scallops & 0.125 & 0.625 & 0.5 \\
breakfast\_burrito & 0.6 & 0.2 & 0.5 \\
sashimi & 0.75 & 0.25 & 1 \\
fish\_and\_chips & 0.72 & 0.27 & 0.63 \\
apple\_pie & 0.333 & 0.555 & 0.666 \\ \bottomrule
\end{tabular}
\caption{Accuracy of fusion classifier and single modality classifiers for some food categories. Here, accuracy is the number of correct top-1 predictions for that class.}
\label{tab:class-analysis}
\vspace{-4mm}
\end{table}

\begin{figure*}[h]
  \centering
  \includegraphics[width=\linewidth]{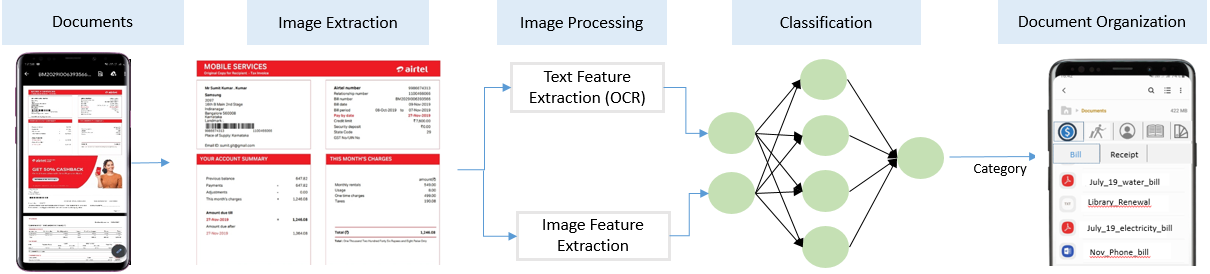}
  \caption{Pipeline of the on-device document classification framework.}
  \Description{Pipeline}
\label{img:pipeline}
\end{figure*}

\section{On-Device Document Classification}
We define On-Device document classification as a task of categorising real documents on user device into topics. These topics were decided based on an internal survey of \textasciitilde 100 people. For this task, we created an in house dataset of \textasciitilde 512 documents as presented in Table \ref{tab:data-stats}. For the ease of purpose, we consider only PDF documents, but the architecture can be easily extended to any other document type(word, excel, etc.). The architecture as shown in Figure \ref{img:pipeline}, is divided into 3 parts: data extraction, feature generation and classification.

\textbf{Data Extraction} : While libraries such as pdfbox exist to extract text from a PDF, it will restrict our document set to only PDFs. To create a framework for processing all document types, we use OCR to extract text from PDF. We convert the PDF documents into a set of images and use MLKit\footnote{\url{https://firebase.google.com/docs/ml-kit/recognize-text }} to extract text from the images. We then use the images and extracted text as input data to our Multimodal Architecture presented in Figure \ref{Model-arch} with some changes to model parameters. For now, we consider only the first page of the PDF document, since the document may consist of 10s of pages and processing all of them would increase latency.

\textbf{Feature Generation} : OCR text extraction is highly erratic for PDF documents since it parses information in a line manner. For example, if a PDF is column split, like research papers, the sequence of text will be lost. Thus, a word order based model may fail for this task. We use regex-based text filtering and the pre-processing approach mentioned in the previous section. Further, to tackle noise due to watermarks, etc., we use a filter to remove non-English words from the tokenized text. We maintain a pre-defined a vocabulary of 60,000 english words to extract Tf-Idf vectors for a document. This vocabulary covers 98\% of words present in our dataset. For image features, we use the image generated while extracting text as input to the visual modality network.

\begin{table}
\begin{tabular}{@{}ll@{}}
\toprule
\textbf{Category} & \textbf{\# of documents} \\ \midrule
Travel & 118 \\
Personal information & 100 \\
Receipts & 98 \\
papers/books & 102 \\
Misc & 94 \\ \bottomrule
\end{tabular}
\caption{Class split of the document dataset.}
\label{tab:data-stats}
\vspace{-8mm}
\end{table}

\textbf{Classification} : We use the architecture presented above for FOOD-101 dataset for this task. We tweak the network parameters to account for the complexity of this specific task and dataset. For the text classification, we reduced the rank of Tf-Idf vector to 200 using SVD. Following which, we trained 2 dense layers of size 64 and 32 units respectively before adding a final softmax layer for classification. For the image classification, we use MobileNet architecture and train a pooling layer of 512 units and a dense layer of 64 units. We followed the same approach of pre-training text and image classifier separately, and trained a fusion classifier on top of it. The results of single modalities, as well as fusion model, are presented in Table \ref{tab:accuracy}.

\begin{figure}
	\begin{subfigure}[h]{0.45\linewidth}
	\includegraphics[width = 1.0\linewidth]{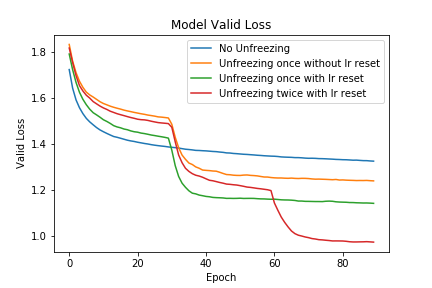}
	\caption{Loss curve of models}
	\end{subfigure}
	\hfill
	\begin{subfigure}[h]{0.45\linewidth}
	\includegraphics[width = 1.0\linewidth]{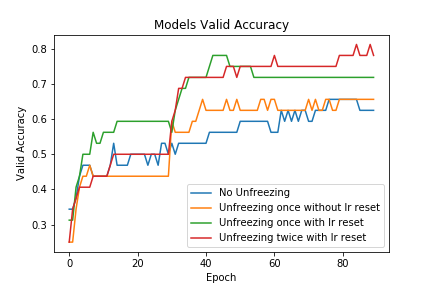}
	\caption{Accuracy curve of models}
	\end{subfigure}
\caption{Comparing validation loss and accuracy of different training techniques}
\label{img:tl-loss-acc}
\vspace{-4mm}
\end{figure}

\textbf{Training Methodology} : We use Stochastic Gradient Descent with decay as our optimiser with an initial learning rate of 0.01 and decay of 0.8 after every epoch. For training fusion classifier, we follow a multi-stage transfer learning approach, inspired by \cite{howard2018universal}. We experiment with various strategies of unfreezing layer weights for training. We train a base model for 90 epochs without unfreezing the layers of the pre-trained models. MobileNet architecture consists of repeated blocks of point-wise and depth-wise separated convolutions. So, we unfreeze weights post the end of a block of MobileNet and at the end of second layer of our text model architecture. The layer numbers in the subsequent section signify the actual layer number in order of our model as available in keras's\cite{chollet2015keras} model summary. Firstly, we unfreeze the weights of pre-trained weights individual modalities models from the 53rd layer of the network after the 30th epoch with and without resetting the learning rate. Secondly, we unfreeze the model weights from 81st layer after the 30th epoch and 53rd layer post the 60th epoch, resetting learning rate at both instances. The accuracy and loss variation is presented in Figure \ref{img:tl-loss-acc}.  Since, our dataset and the dataset on which mobilenet is trained is dissimilar, we observed better performance by resetting the learning rate in stages.

\begin{table}[]
\begin{tabular}{@{}ll@{}}
\toprule
\textbf{Modality} & \textbf{Accuracy} \\ \midrule
Text & 59.62\% \\
Visual & 71.88\% \\
Text + Visual & 84.38\% \\ \bottomrule
\end{tabular}
\caption{Top-1 Accuracy of single and multimodal models.}
\label{tab:accuracy}
\vspace{-8mm}
\end{table}

\textbf{On-Device Execution}
Depending upon the extension of a file, we can choose an appropriate renderer. For our pdf documents, we use android PdfRenderer\footnote{\url{https://developer.android.com/reference/android/graphics/pdf/PdfRenderer}} for rendering. Once the rendering is done, we extract bitmap which is processed to get Image features. To extract the text of the document, we have used Google Mlkit\footnote{\url{https://firebase.google.com/docs/ml-kit/recognize-text}} which provides on-device API for text extraction. For simplicity, only the English language is considered. However, it's possible to extract text for different languages on-device as explained in \cite{kumar2020device}. The offline fusion model which was trained, is quantized using the tflite\footnote{\url{https://tensorflow.org/lite}} post-training quantization method. The quanitzation led to an accuracy reduction of \textasciitilde 0.5\%, a minor impact considering it lead to model compression of 75\%. The size of the final model is \textasciitilde 13 MB. The total execution time of this pipeline on a document is 4.6s, out of which ~3.5s is taken by OCR.

\section{Conclusion}
The storage capacity of smartphones is ever-increasing, which leads to a vast accumulation of documents on-device. Such a clutter inhibits a user from retrieving relevant documents. Moreover, with the internet becoming increasingly multimodal, we should leverage the information offered by different modalities for a better understanding of content. With this work, we show that different modalities indeed contribute towards increased understanding of documents. We achieve a \textasciitilde 90\% accuracy with our fusion network on FOOD 101 dataset, matching the previous best with a reduction in model size. We also present a feasible framework to classify documents on-device. We acknowledge that the size of our dataset is not sufficient and conclusive evidence of the same. But we hope that our work serves as a precursor for others to contribute to this field of on-device document classification. 

\bibliographystyle{ACM-Reference-Format}
\bibliography{references}
\end{document}